\documentclass[10pt,twocolumn,letterpaper]{article}

\usepackage[pagenumbers]{cvpr} 
\usepackage{multirow}
\usepackage{makecell}
\usepackage{amsmath}
\usepackage{algorithm}
\usepackage{graphicx}
\usepackage{algorithmic}
\usepackage[accsupp]{axessibility}

\usepackage[dvipsnames]{xcolor}

\definecolor{cvprblue}{rgb}{0.21,0.49,0.74}
\usepackage[pagebackref,breaklinks,colorlinks,citecolor=cvprblue]{hyperref}

\def\paperID{2} 
\def\confName{CVFAD}
\def\confYear{2024}

\title{Attribute-Guided Multi-Level Attention Network \\for Fine-Grained Fashion Retrieval}

\author{Ling Xiao \\
The University of Tokyo\\
{\tt\small ling@cvm.t.u-tokyo.ac.jp}
\and
Toshihiko Yamasaki\\
The University of Tokyo\\
{\tt\small  yamasaki@cvm.t.u-tokyo.ac.jp}
}

\begin{document}

\maketitle
\begin{abstract}
Fine-grained fashion retrieval searches for items that share a similar attribute with the query image. Most existing methods use a pre-trained feature extractor (e.g., ResNet 50) to capture image representations. However, a pre-trained feature backbone is typically trained for image classification and object detection, which are fundamentally different tasks from fine-grained fashion retrieval. Therefore, existing methods suffer from a feature gap problem when directly using the pre-trained backbone for fine-tuning. To solve this problem, we introduce an attribute-guided multi-level attention network (AG-MAN). Specifically, we first enhance the pre-trained feature extractor to capture multi-level image embedding, thereby enriching the low-level features within these representations. Then, we propose a classification scheme where images with the same attribute, albeit with different values, are categorized into the same class. This can further alleviate the feature gap problem by perturbing object-centric feature learning. Moreover, we propose an improved attribute-guided attention module for extracting more accurate attribute-specific representations. Our model consistently outperforms existing attention based methods when assessed on the FashionAI (62.8788\% in MAP), DeepFashion (8.9804\% in MAP), and Zappos50k datasets (93.32\% in Prediction accuracy). Especially, ours improves the most typical ASENet\_V2 model by 2.12\%, 0.31\%, and 0.78\% points in FashionAI, DeepFashion, and Zappos50k datasets, respectively. The source code is available in \href{https://github.com/Dr-LingXiao/AG-MAN}{https://github.com/Dr-LingXiao/AG-MAN}.
\end{abstract}

\section{Introduction}
\label{sec:introduction}
Fashion retrieval is a crucial research topic~\cite{Huang_ICCV_15, Liu_CVPR_16,Liang_IEEE-TMM_16,Chen_IEEE-TMM_17,Gajic_CVPR_18,Gu_IEEE-TMM_18,Ak_WACV_18,Kuang_ICCV_19,Jing_IEEE-TMM_19,Jing_IEEE-TMM_21, Liu_MM_21,Baldrati_ACMMM-Asia_21, Ning_CIS_22,Dodds_arxiv_22,Goenka_CVPR_22,Xiao_ICIP_22,Liao_ACMMM_18}, especially for fashion recommendations~\cite{Zhang_IEEE-TMM_17,Liu_IEEE-TMM_20,Zhan_IEEE-TMM_21,Ding_IEEE-TMM_21,Lu_IEEE-TMM_22}. It aims to learn the overall similarity among different fashion items~\cite{Zhao_CVPR_17,Han_ACMMultimedia_17, Gu_IEEE-TMM_18}. Fine-grained fashion retrieval is an emerging topic in the fashion retrieval field. It focuses on learning fine-grained similarity rather than overall similarity, because the similarity between items varies depending on the specific attribute considered in the task. For example, two shirts with a crew neck and a v-neck, respectively, are considered similar in a fashion retrieval task, while they will be treated as different items when doing fine-grained fashion retrieval under the guidance of neckline design. Moreover, the attribute can be very diverse, thus this research task is more challenging than the general fashion retrieval.

This research topic, while challenging, holds significant importance in the field. Achieving high-quality fine-grained fashion retrieval is crucial not only for meeting the varied and diverse needs of users in the real world but also plays a pivotal role in fashion copyright protection, as highlighted by Martin {\it et al.}~\cite{Martin_U_19}. This capability is instrumental in identifying items with copied or duplicated designs, which hinges on the nuanced ability to discern intricate similarities among fashion items. The process involves analyzing detailed attributes of clothing, such as texture, pattern, and shape, at a granular level, enabling the detection of subtle differences and similarities that are not apparent at a glance. As the fashion industry continues to grow and evolve, with an ever-increasing volume of designs and the rapid proliferation of knock-offs, the demand for advanced fine-grained similarity analysis technologies becomes more pressing. These technologies not only support the creative integrity of designers by safeguarding their original designs but also enhance consumer experience by providing more accurate and personalized fashion recommendations. Therefore, developing and refining these technologies is essential for advancing both copyright protection efforts and the overall user experience in fashion e-commerce platforms.


Current approaches for fine-grained fashion retrieval are mainly the conditional similarity network~\cite{Veit_CVPR_17} and attention networks~\cite{Ma_AAAI_20,Wan_ETAI_22,Yan_ICME_22}. The most recent contrastive learning method~\cite{xiao_MIPR_23} and clustering method~\cite{Jiao_ECCV_22} use the attention network~\cite{Ma_AAAI_20} as their base network.

However, existing state-of-the-art (SOTA) methods have two problems. (1) They rely on pre-trained Convolutional Neural Network (CNN) backbones that were initially trained for image classification on the ImageNet dataset~\cite{Deng_CVPR_09} to extract image representations. The captured image embedding is then processed to calculate the fine-grained similarities. This leads to a feature gap problem due to the distinct nature of the image classification task and fine-grained fashion retrieval task. (2) Existing work adopts high level features for fine-grained fashion similarity learning. Considering the diversity of the attributes, the neglect of low level features will degrade the model performance, especially for some attributes that care about small texture difference.

To address the above-mentioned issue, we present an attribute-guided multi-level attention network (AG-MAN) to improve retrieval accuracy in fine-grained fashion similarity learning. The proposed AG-MAN can extract more discriminative image features. Specifically, we firstly enhance the pre-trained CNN backbone to increase the low-level features contained in image representations. Then, when fine-tuning the pre-trained CNN for extracting image features, we suggest incorporating a classification loss that groups images sharing the same attribute but differing in sub-classes into a common category. This can further alleviate the feature gap problem by perturbing for object-centric feature learning. Once improved image representations are attained, we introduce an improved attribute-guided attention module to derive more accurate attribute-specific representations. The proposed AG-MAN consistently outperforms existing attention networks over three datasets. In brief, this paper offers the following key contributions:

\begin{enumerate}
\item{This paper identifies the feature gap problem in SOTA fine-grained fashion retrieval methods and proposes solutions to address it.} 

\item{We propose an improved attribute-guided attention module, named attribute-guided attention (AGA).}

\item{The proposed attribute-guided multi-level attention network, AG-MAN, consistently outperforms existing methods across three datasets}. 
\end{enumerate}

\begin{figure*}[t]
  \centering
   \includegraphics[width=0.9\linewidth]{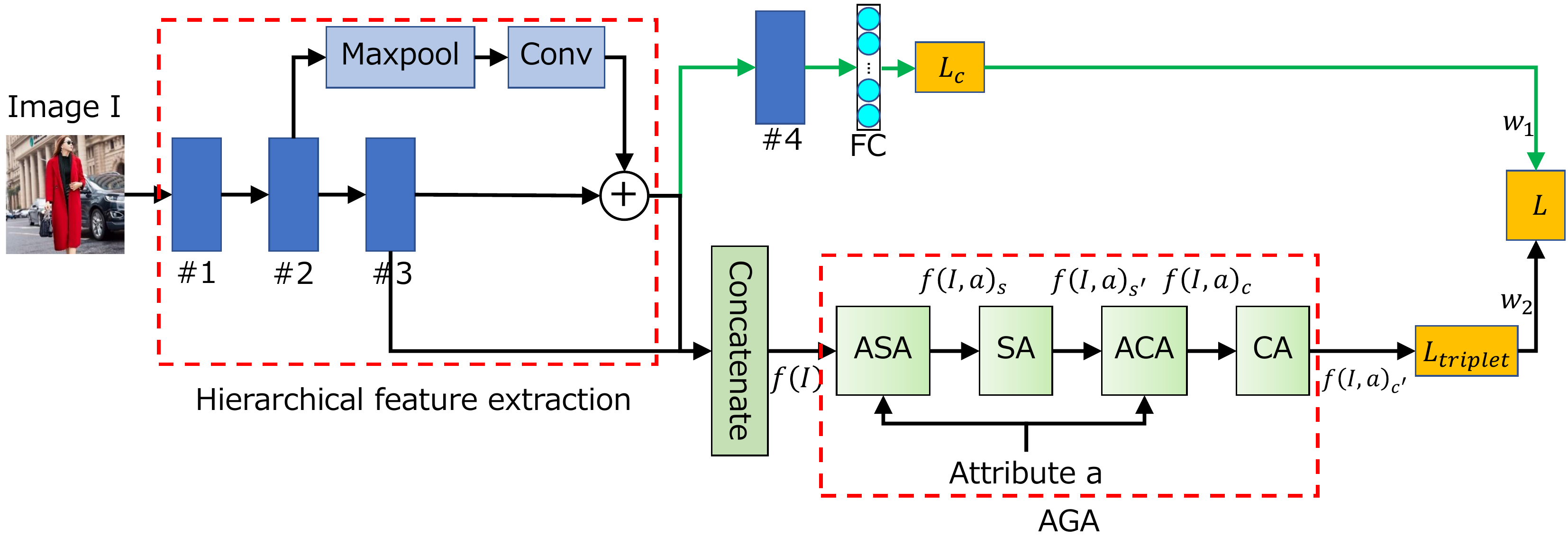}
   \caption{Model architecture. \#1, \#2, \#3, and \#4 are four blocks in ResNet50 backbone.}
   \label{fig:model}
    \vspace{-0.2cm}
\end{figure*}

\begin{figure}[t]
  \centering
   \includegraphics[width=1.0\linewidth]{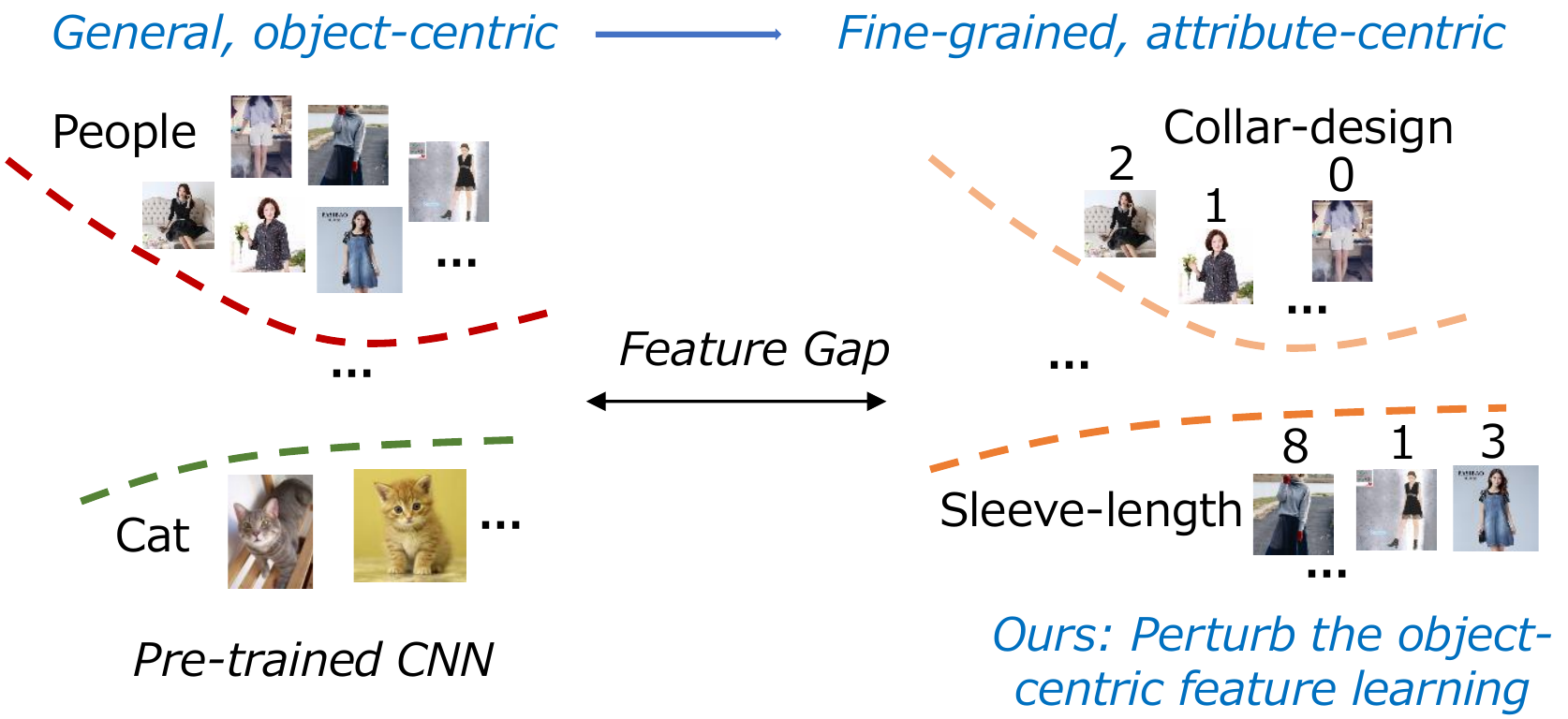}
   \caption{The proposed classification branch perturbs the object-centric feature learning in the fine-tuning process by grouping images with the same attribute but different sub-classes into the same class. The number above the image denotes the sub-class under each attribute.}
   \label{fig:classification}
   \vspace{-0.2cm}
\end{figure}

\section{Proposed method}

\subsection{Overview}

The primary motivation of this research is to tackle the feature gap issue that currently challenges SOTA attention networks in fine-grained fashion retrieval. This problem arises from the common practice of directly employing pre-trained CNN backbones to extract image representations, which are subsequently processed with attention modules. Our underlying theoretical foundation stems from the notion that leveraging a feature embedding for classification may hinder the learning process for the given task, mainly because it overlooks essential lower-level image features. Therefore, it is imperative to exercise caution when employing pre-trained CNNs, especially when the new research task diverges significantly from traditional image classification.

Figure~\ref{fig:model} depicts the architecture of our proposed network. In this research topic, the query attribute can be related to the local information (e.g., neckline-design, texture-related, and lapel-design, etc.) and global information (e.g., style-related, pant-related, and shape-related, etc.). Therefore, capturing multi-level image features is crucial rather than solely focusing on designing complex attention modules. Thus, we first enhance the pre-trained CNN backbone to capture hierarchical image features. We also propose to calculate the logit distribution difference between images with different attribute labels by adding one classification loss $L_{c}$. This can disrupt the object-centric feature learning within a pre-trained CNN backbone during the fine-tuning process, thus contributing to the mitigation of the feature gap issue. To extract more accurate attribute-specific embeddings, we introduce an improved attribute-guided attention (AGA) module. 

Concretely, for a given query item $(I, a)$, with $I$ denoting the image and $a$ indicating the attribute, the attribute attended feature vector is $f(I, a)\in \mathbb{R}^c$. This feature vector lies in a continuous vector space $\mathbb{R}^c$ and encapsulates the distinctive features of the specified attribute within the image. Here, $c$ represents the dimensionality of the feature vector. For a pair of fashion images, $(I, I^{'})$, we compute the embedding similarity between their respective attribute-specific feature vectors, $(f(I, a),~f(I^{'}, a))$, to quantify their similarity with respect to the same attribute. Furthermore, we extend this similarity assessment across multiple attributes by aggregating the similarity scores for each individual attribute. Specifically, we calculate the similarity score for each attribute separately and then sum these scores to obtain an overall measure of fine-grained similarity between the two images across all attributes. It is essential to emphasize that each attribute-specific feature vector resides in its respective attribute-specific embedding space. If there are a total of $n$ attributes of interest, we learn $n$ separate embedding spaces simultaneously to capture the distinct characteristics of each attribute within the image.

\begin{figure*}[t]
  \centering
   \includegraphics[width=1.0\linewidth]{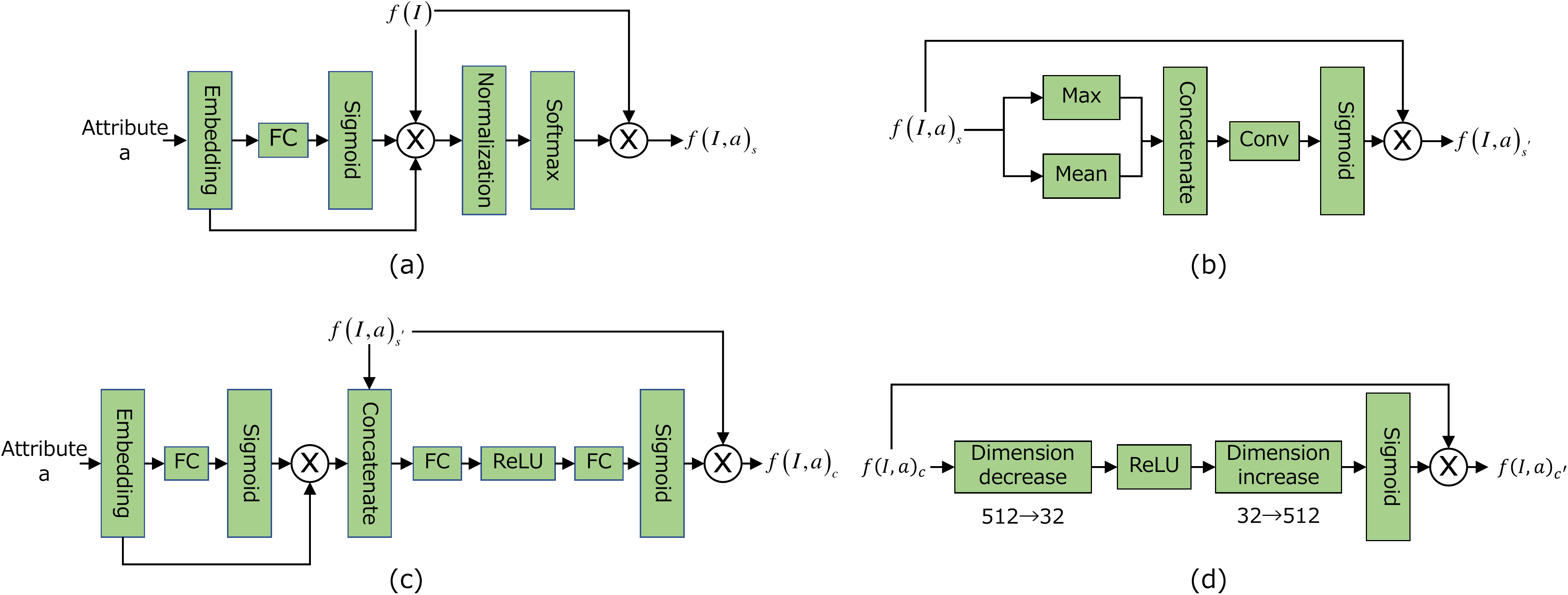}
   \caption{Details of (a) ASA, (b) SA, (c) ACA, and (d) CA in the proposed AGA module. The ASA and SA enhance the related region localization while the ACA and CA improve the ability to distinguish between different attributes within the same region.}
   \label{fig:AGA}
\end{figure*}

\subsection{Hierarchical feature extraction}

The concept of hierarchical features has been widely applied across various research domains. Yet, there is a lack of research focused on designing an effective hierarchical feature extraction module specifically for fine-grained fashion retrieval. In this paper, we propose to develop a hierarchical feature extraction module tailored for fine-grained fashion retrieval. We use the widely adopted pre-trained ResNet50~\cite{He_CVPR_16} as the backbone network for a fair comparison with exsiting models\cite{Veit_CVPR_17,Ma_AAAI_20,Wan_ETAI_22}. Because fine-grained fashion retrieval needs to deal with multiple attributes, and it cares about different level features, integrating more multi-level features into the image features is essential. To achieve this, we begin by performing an addition operation to combine the $F_{2}(.)$ (output of intermediate block~2 of ResNet50) and $F_{3}(.)$, resulting in a fused output denoted as $F_{2,3}(.)$. Next, we concatenate $F_{2,3}(.)$ with $F_{3}(.)$ and feed this concatenated feature representation into the proposed attribute-guided attention module. This can enhance the multi-level features contained in the image representations that will be processed with the attribute-guided attention module.

Moreover, the pre-trained CNN is designed to learn some object-centric features, which are not what we need in fine-grained fashion retrieval. Thus, we propose adding a classification loss to group images with the same attribute but different sub-classes into the same class (Figure~\ref{fig:classification}). This can boost the learning of more diverse image features. Concretely, the $F_{2,3}(.)$ is passed into block~4 of ResNet50~\cite{He_CVPR_16} and a fully connected (FC) layer to calculate a classification loss of different attributes.

\subsection{AGA}

The proposed AGA comprises four parts: attribute-guided spatial attention (ASA), spatial attention (SA), attribute-guided channel attention (ACA), and channel attention (CA). It is an improvement of the attribute-aware attention module proposed in~\cite{Ma_AAAI_20}. To encode the attribute information, we use a one-hot vector denoted as $a\in \{0, 1\}^{n}$, where $n\in N$ represents the total number of distinct attributes.

\textbf{\noindent ASA}
In this research task, learning feature representations for specific, related regions is essential. For instance, to capture attribute-specific features for the neck design, the region surrounding the neck holds greater significance compared to others regions. Moreover, since fashion images can exhibit substantial variations, such as in poses and scales, relying on a fixed region for a specific attribute across all images is suboptimal. Thus, this paper introduce an ASA module, a mechanism that dynamically focuses on specific areas of an input image based on the relevance with a particular attribute, as illustrated in Figure~\ref{fig:AGA}a. To elaborate, with $h$ and $w$ representing the height and width dimensions, we first align the dimensionality of the image feature embedding $f(I) \in \mathbb{R}^{c\times h\times w}$ and the attribute embedding $p(a)\in \mathbb{R}^{c^{'}\times h\times w}$. Formally, we obtain the transformed image feature representation $p(I) \in \mathbb{R}^{c^{'}\times h\times w}$ by applying a convolutional layer $\mathrm{Conv}_{c^{'}}$ comprising $c^{'}$ $1\times 1$ convolution kernels to the $f(I) \in \mathbb{R}^{c\times h\times w}$, given as
\begin{equation}
  p(I) = \tanh(\mathrm{Conv}_{c^{'}}(I)).
  \label{eq:P-I}
\end{equation}

To obtain attribute representations, we initially transform the query attribute into a vector of dimensionality $c^{'}$, which is achieved via an embedding layer and an FC layer. Subsequently, we apply self-attention to the attribute embedding. Therefore, the transformed attribute $p(a)$ belongs to the space $\mathbb{R}^{c^{'}\times h\times w}$, where $c'$ denoting the number of channels, and 
\begin{equation}
  p(a) = \sigma(W_{a}a)*W_{a}a,
  \label{eq:P-a}
\end{equation}
where $W_{a}\in \mathbb{R}^{c^{'}\times n}$ represents the transformation matrix for the embedding and FC layers, $n$ represents the number of attributes, and $\sigma$ indicates a sigmoid function. Finally, the spatially attended representation of the image is given by
 \begin{equation}
  f(I, a)_{s} = f(I) * W_{1}s(p(I)\cdot p(a)),
  \label{eq:I-s}
\end{equation}
where $s$ denotes a softmax function and $W_{1}$ indicates a learned weight matrix. It's important to highlight that we use the sigmoid function for optimizing the attention weights in most cases, a common practice in many attention methods, as described in~\cite{Woo_ECCV_18,Hu_CVPR_18}. In Eq.~\ref{eq:I-s}, the softmax is adopted to further improve the model's ability in distinguishing the differences between features of different attributes.



\textbf{\noindent SA}
After obtaining $f(I, a)_{s}$, we first calculate the average and max channel pooling outputs of $f(I, a)_{s}$, denoted as $\mathrm{Avg}(I_{s})$ and $\mathrm{Max}(I_{s})$, and concatenate them in the channel dimension to retain more information. They are then passed into a convolution layer and a sigmoid layer to extract an attention weight for the $f(I, a)_{s}$. SA aims to further improve the positioning ability of $f(I, a)_{s}$. The processed $f(I,a)_{s}$ is denoted as $f(I,a)_{s^{'}}$. The detailed operation is given below:

\begin{subequations}\label{eq:I-ss}
\begin{align}
W_{SA}&= \sigma(\mathrm{Conv}_{1}([\mathrm{Avg}(I_{s}), \mathrm{Max}(I_{s})])), \label{eq:I-ss-A}\\
f(I, a)_{s^{'}} &= f(I, a)_{s} * W_{SA}, \label{I-ss-B}
\end{align}
\end{subequations}
where $\mathrm{Conv}_{1}$ represents a convolutional layer consisting of a single $1\times1$ convolution kernel and $[,]$ denotes a concatenation operation. 

The proposed SA is very straightforward. We concatenate the $\mathrm{Avg}(I_{s})$ and $\mathrm{Max}(I_{s})$ for more diverse information (Figure~\ref{fig:AGA}b).

\textbf{\noindent ACA}
While the ASA and SA dynamically concentrate on particular regions within the image, it's important to note that these identical regions might simultaneously relate to several attributes simultaneously. For instance, attributes such as lapel design and neck design are both associated with the area surrounding the neck. To address this issue, we introduce an ACA mechanism applied to the spatially attended embedding, denoted as $f(I, a)_{s^{'}}$. The ACA is designed to select dimensions within the spatially attended feature vector that are relevant to the given attribute. In more detail, we start by transforming the attribute into an embedding vector with an embedding layer and a FC layer. This embedding vector has the same dimensionality as $f(I, a)_{s^{'}}$, and is expressed as

 \begin{equation}
  q(a) = \sigma(W_{a^{'}}a)*(W_{a^{'}}a),
  \label{eq:q-a}
\end{equation}
where $W_{a^{'}} \in \mathbb{R}^{c^{'} \times n}$ is a transformation matrix. 

Then, the attribute representation $q(a)$ and spatially attended image embedding $f(I, a)_{s^{'}}$ are combined by an element-wise addition and the resultant feature is passed through two subsequent FC layers to compute the channel attention weights. We also adopt a learned weight matrix $W_{2}$ to increase the adaptability. The final output is

\begin{subequations}\label{eq:I-c}
\begin{align}
W_{ACA} &= \sigma(W_{c_{2}}(r(W_{c_{1}}([q(a), f(I, a)_{s^{'}}])))), \label{eq:I-c-A}\\
f(I, a)_{c} &= f(I, a)_{s^{'}} * W_{2}*W_{ACA}, \label{I-c-B}
\end{align}
\end{subequations}
where $W_{c_{1}}$ and $W_{c_{2}}$ indicate the weight matrices of two FC layers, respectively, and $r$ represents a ReLU activation function.

\textbf{\noindent CA}
To enhance the utilization of pertinent channel information derived from the obtained $f(I, a)_{c}$, we implement a simple dimensionality reduction, succeeded by an expansion operation, aimed at extracting channel-specific attention. This process yields $f(I, a)_{c^{'}}$. For a more detailed description, please refer to Figure~\ref{fig:AGA}d.

\begin{table*}[t]
\caption{Details of used datasets. They provide a rich set of attributes and their corresponding sub-classes to describe fashion items in detail.}
\begin{center}
\resizebox{0.75\textwidth}{!}{
\begin{tabular}{lclc}
\toprule
Dataset &Attribute& Sub-classes & Total\\
\cline{1-4}
\multirow{8}{*}{\textbf{FashionAI}}&skirt-length & short length, knee length, midi length, etc. &6   \\ 
  &sleeve-length &sleeveless, cup sleeves, short sleeves, etc.&  9\\
  &coat-length &high waist length, regular length, long length, etc.  & 8 
  \\
  &pant-length &short pant, mid-length, 3/4 length, etc. &6  \\
  &collar-design & shirt collar, peter pan, puritan collar, etc. & 5\\
  &lapel-design & notched, collarless, shawl collar, etc. & 5\\
  & neckline-design &strapless neck, deep V neckline, straight neck, etc. & 11 \\
  &neck-design & invisible, turtle neck, ruffle semi-high collar, etc. & 5\\ \midrule
\multirow{5}{*}{\textbf{DeepFashion}}&texture-related & abstract, animal, bandana, etc. & 156 \\
&fabric-related & acid, applique, bead, etc. & 218 \\
&shape-related & a-line, ankle, asymmetric, etc. & 180 \\
&part-related & arrow collar, back bow, batwing, etc. & 216  \\
& style-related & americana, art, athletic, etc. &230 \\ \midrule
\multirow{4}{*}{\textbf{Zappos50k}}&category & shoes, boots, sandals, etc. & 4 \\
 & gender & women, men, girls, etc. & 4 \\
 & heel height & 1-4in, 5in\&over, flat, etc. & 7\\
 &closure & buckle, pull on, slip on, etc.  &19 \\ \bottomrule
\end{tabular}}
\end{center}
\label{tab:dataset}
\end{table*}

\begin{figure*}[t]
  \centering
   \includegraphics[width=0.65\linewidth]{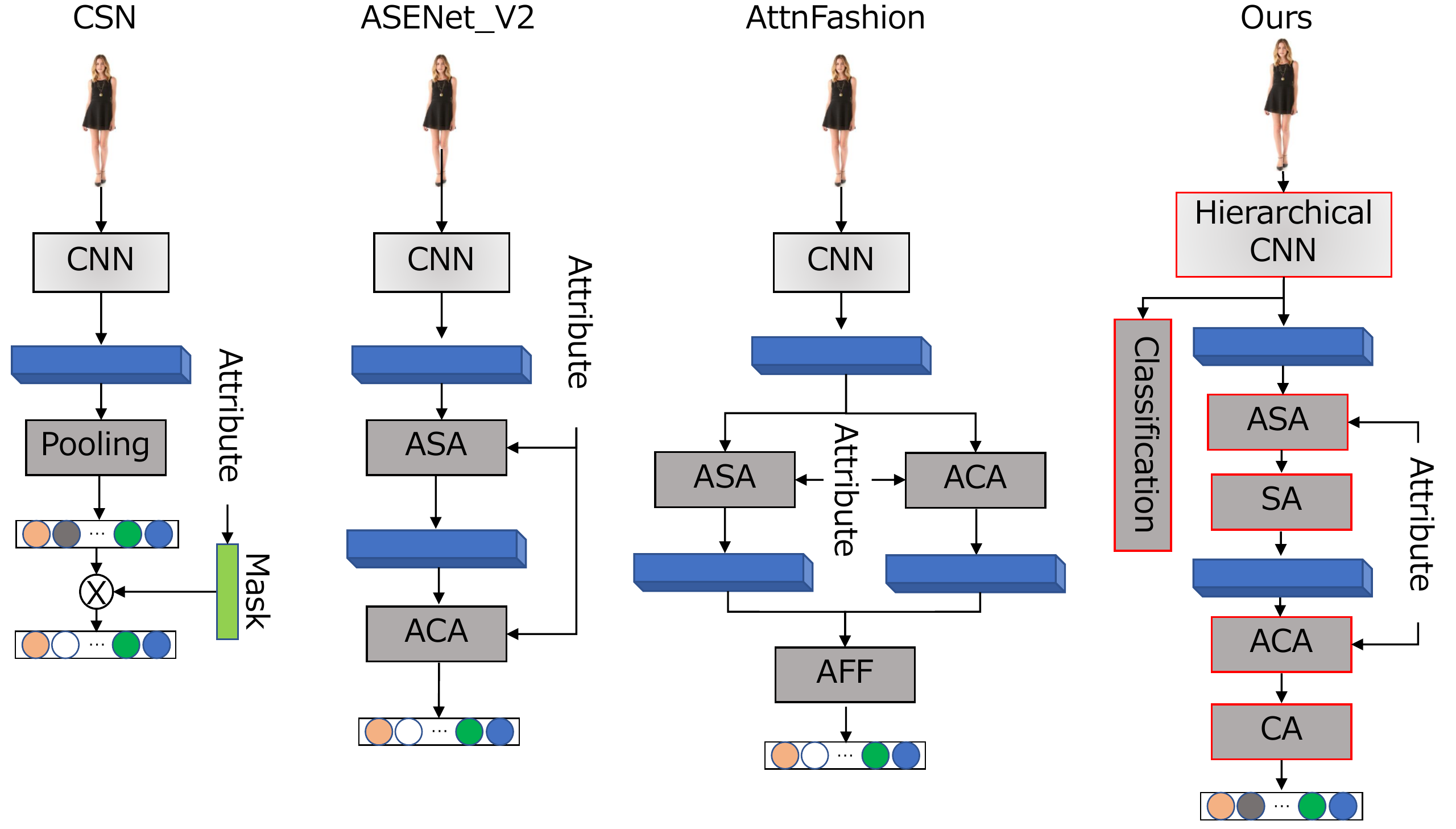}
   \caption{The conceptual structure of compared models and ours. The CSN is a conditional similarity network while the the ASENet\_V2 and AttnFashion are attention networks in this research field. Ours is also an attention network.}
   \label{fig:compared_models}
\end{figure*}

\begin{table*}[t]
\small
\caption{Results on the FashionAI dataset: The best result is bolded, and the second best is underlined. Our model consistently surpasses its counterparts across all attribute types in terms of performance.}
\begin{center}
\resizebox{1.0\textwidth}{!}{
\begin{tabular}{lccccccccc}
\toprule
\multirow{2}{*}{Methods}& \multicolumn{8}{c}{MAP for each attribute (Fashion AI)}& \multirow{2}{*}{MAP $\uparrow$}\\
\cline{2-9}
   &Skirt-length & Sleeve-length & Coat-length & Pant-length &Collar-design &Lapel-design &Neckline-design &Neck-design & \\
\midrule
CSN~\cite{Veit_CVPR_17} & 59.5852& 43.9943& 42.8203& 62.8021&67.5715 &48.1033 & 41.7291& 55.2886& 50.8043 \\ \midrule
 ASENet\_V2~\cite{Ma_AAAI_20} &\underline{64.5731} &\underline{54.9630} &\underline{51.7645} & \underline{64.4973} &\underline{71.9256}& \underline{66.7201}& \underline{60.2929} &\underline{60.8307}&\underline{60.7551}  \\ \midrule
 AttnFashion~\cite{Wan_ETAI_22} & 61.9057&45.8526 &45.8116 & 60.6013&70.4638 &62.7362 &52.0484 &59.5861 &55.3646   \\ \midrule
 Ours &\textbf{65.5674}  & \textbf{55.0840} &  \textbf{55.2060}& \textbf{68.9536} &\textbf{73.7867} &\textbf{68.1758}   &\textbf{63.1985}   &\textbf{62.0777}   & \textbf{62.8788}  \\ \bottomrule
\end{tabular}}
\end{center}
\label{tab:Main_results_FashionAI}
 \vspace{-0.2cm}
\end{table*}

\begin{table*}[t]
\caption{Experimental results on DeepFashion dataset. Our model performs the best.}
\begin{center}
\resizebox{0.8\textwidth}{!}{
\begin{tabular}{lcccccc}
\toprule
\multirow{2}{*}{Methods}& \multicolumn{5}{c}{MAP for each attribute (DeepFashion)} & \multirow{2}{*}{MAP $\uparrow$}\\
\cline{2-6}
  &Texture-related & Fabric-related & Shape-related & Pant-related
 &Style-related & \\
\midrule
 CSN~\cite{Veit_CVPR_17}& 14.4546 & 6.5021& 11.2555& 4.7751& 3.4673&8.0747   \\ \midrule
  ASENet\_V2~\cite{Ma_AAAI_20} &\underline{15.5168}
  &\underline{7.1855} &\underline{11.5066} &\underline{5.5220} &\underline{3.6514} &  \underline{8.6704}\\ \midrule
 AttnFashion~\cite{Wan_ETAI_22} &11.8423  &5.7248 &8.6115 &4.0904 &3.2329 & 6.8060 \\ \midrule
Ours &  \textbf{15.6624} & \textbf{7.2746} & \textbf{12.3805} &\textbf{5.8935} & \textbf{3.7555} &\textbf{8.9804} \\
 \bottomrule
\end{tabular}}
\end{center}
\label{tab:Main_results_DeepFashion}
 \vspace{-0.2cm}
\end{table*}

\subsection{Model learning}
The model learning process contains two losses: attribute classification loss and triplet ranking loss. First, the attribute classification loss is computed with
 \begin{equation}
L_{c} =-w_{a}[y_{a}\cdot \log\sigma(x_{a})+(1-y_{a})\cdot \log(1-\sigma(x_{a}))].
  \label{eq:classification-loss}
\end{equation}

Then, with the triplet input $(I, I^{p}, I^{n})$, we aim to minimize the distance between the embeddings of images with the same specific sub-class, denoted as $D_{I,I^{p}}$, and maximize the distance between the embeddings of images with different sub-classes, denoted as $D_{I,I^{n}}$. More formally, we define the triplet ranking loss as

\begin{subequations}\label{eq:triplet-loss}
\begin{align}
D_{I,I^{p}} &= \frac{f(I, a)_{c^{'}} \cdot f(I^{p}, a)_{c^{'}}}{\|f(I, a)_{c^{'}}\| \|f(I^{p}, a)_{c^{'}}\|}, \label{eq:eq:triplet-loss-A}\\
D_{I,I^{n}} &=\frac{f(I, a)_{c^{'}} \cdot f(I^{n}, a)_{c^{'}}}{\|f(I, a)_{c^{'}}\| \|f(I^{n}, a)_{c^{'}}\|}, \label{eq:triplet-loss-B} \\
L_{\text{triplet}} &= \max\{0, m + D_{I,I^{p}}- D_{I,I^{n}}\},  \label{eq:triplet-loss-C}
\end{align}
\end{subequations}
where $m$ represents the margin, empirically set to be 0.2.

The final loss is a dynamic weighted loss that combines the attribute classification loss $L_{c}$ and the triplet ranking loss $L_{\text{triplet}}$ in an adaptive manner. The primary goal of defining weighted loss is to control the training process and ensure that both losses contribute effectively to learning. Initially, the $L_{c}$ and $L_{\text{triplet}}$ are set to have the same weight coefficient of $0.5$. Then they will change following an exponential function. To preventing the weight coefficient from becoming too large, which could lead to convergence issues, we add the $w_{0}$ and $w_{1}$ into the weighted loss. Thus, the final loss becomes
 \begin{equation}
L = 0.5*e^{w_{0}}L_{c} +w_{0} + 0.5*e^{w_{1}}L_{\text{triplet}}+w_{1},
  \label{eq:loss}
\end{equation}
where $w_{0}$ and $w_{1}$ are learned weight parameters that are initialized as zero.

\begin{figure*}[t]
  \centering
   \includegraphics[width=0.90\linewidth]{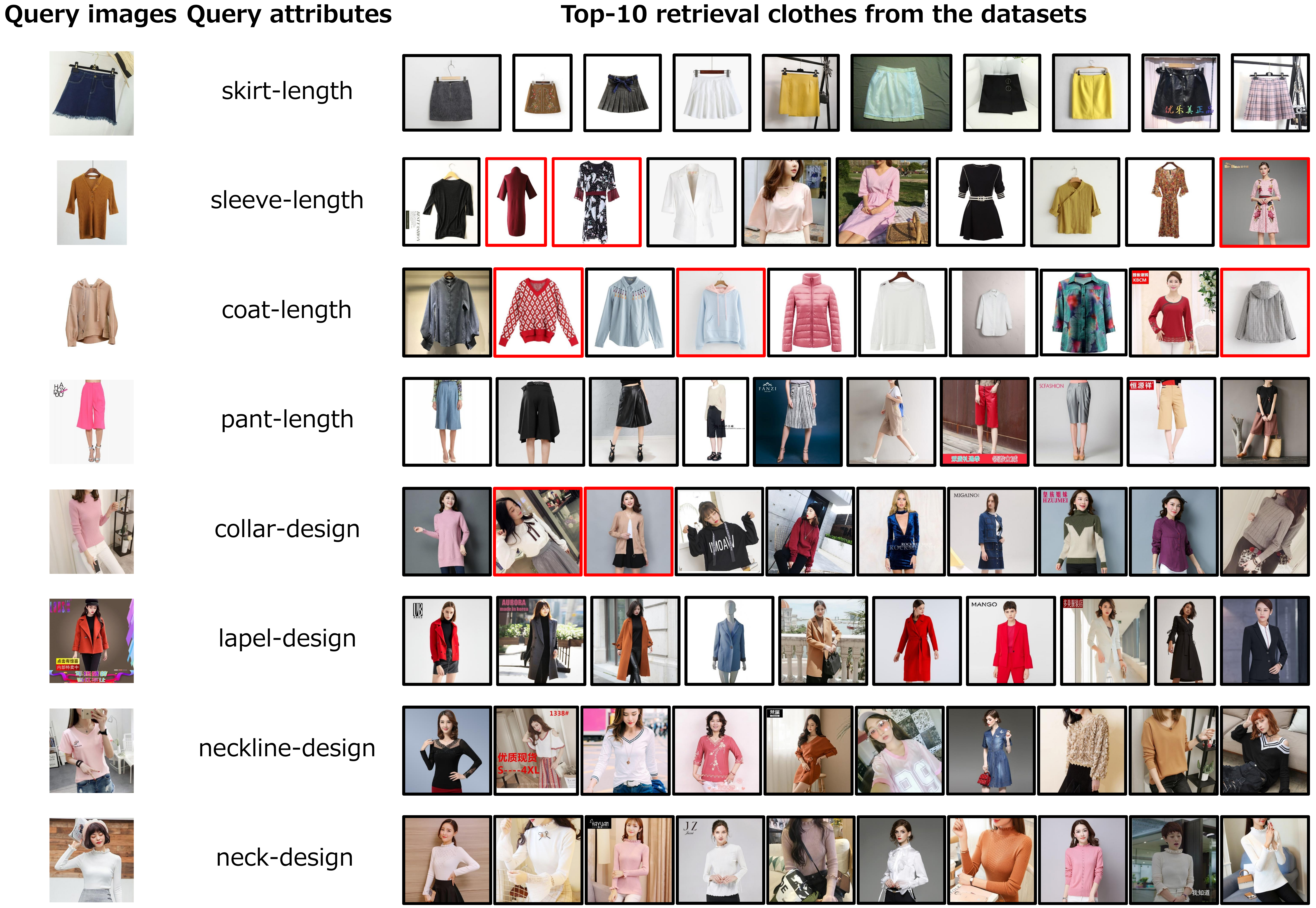}
   \caption{Top-10 retrieval examples from the FashionAI dataset, where the images with a black bounding box exhibit the same sub-classes  with the query image, while images with a red bounding box have different sub-classes compared to the query image. The majority of the search results within the top 10 generated by our model are precise and correct.}
   \label{fig:FashionAI_retri}
\end{figure*}

\begin{figure*}[t]
  \centering
   \includegraphics[width=0.7\linewidth]{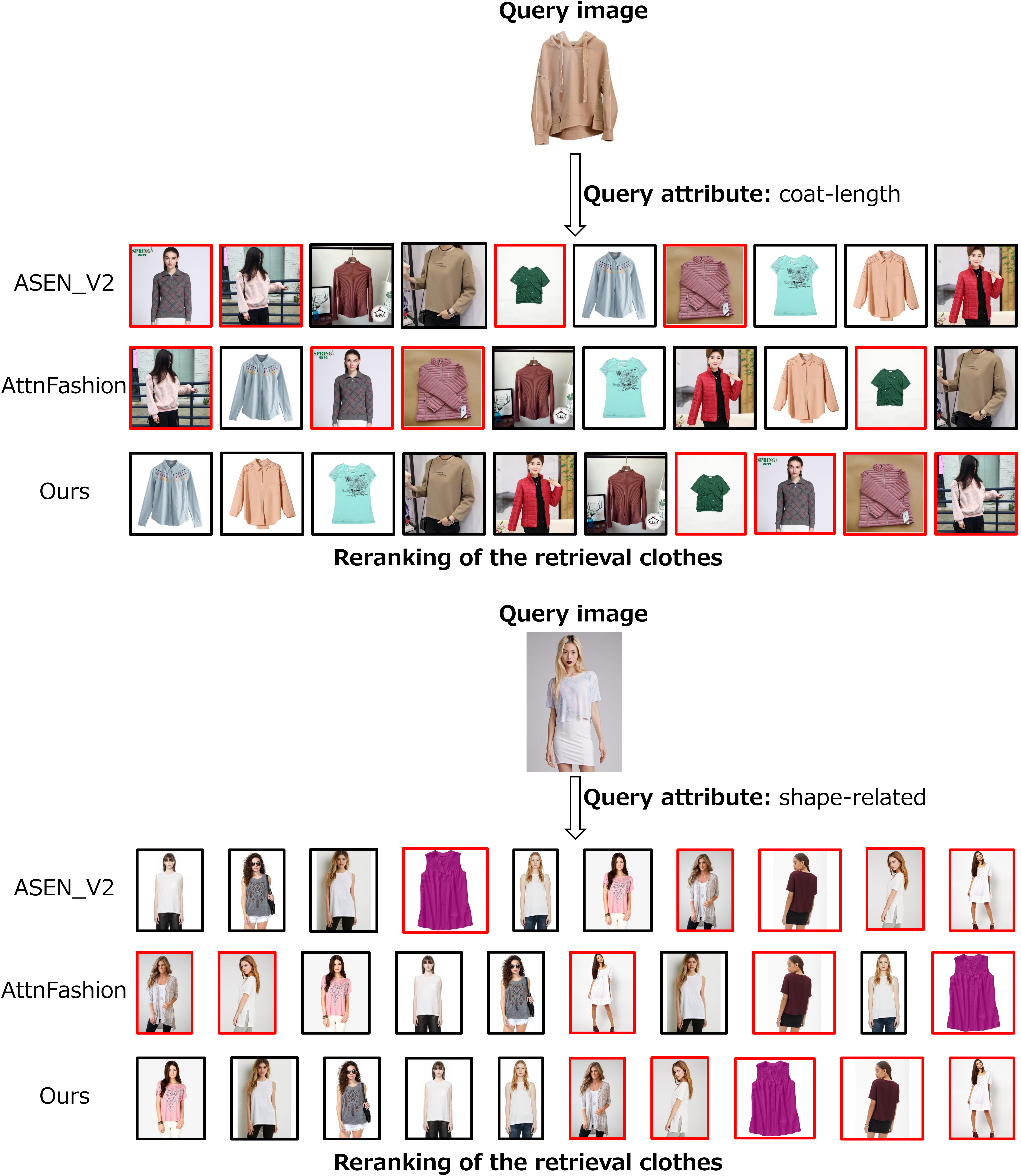}
   \caption{The model ranks the candidates based on their similarity to the query image in terms of the specified attribute. The proposed model effectively distinguishes between candidates that share the same sub-classes as the query image (indicated by a black bounding box) and those that differ (marked with a red bounding box). In contrast, other models struggle to achieve this precise ranking, often placing candidates with shared sub-classes lower in the list than they should.}
   \label{fig:ranking}
    \vspace{-0.2cm}
\end{figure*}

\begin{figure*}[t]
  \centering
   \includegraphics[width=0.95\linewidth]{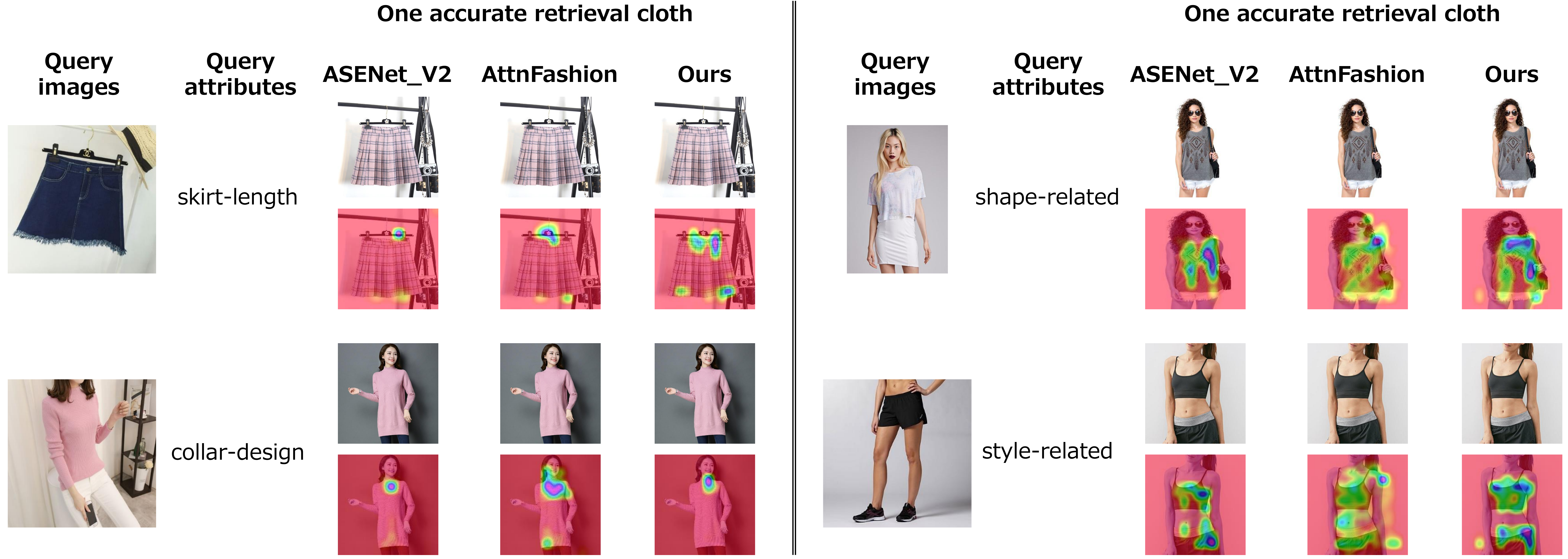}
   \caption{Visualization of the spatial attention based on a specified query attribute, depicted above the original retrieved image.}
   \label{fig:Attenmap}
    \vspace{-0.2cm}
\end{figure*}

\subsection{Material}
A summary of the datasets used in this paper are given in Table~\ref{tab:dataset}.

\textbf{FashionAI.}
FashionAI is a large-scale fashion dataset with hierarchical attribute annotations for fashion understanding. We use this dataset because of its high-quality attribute annotations. Since the full version of FashionAI has not been publicly released, we use an early version as in~\cite{Ma_AAAI_20}. It consists of 180,335 apparel images, where each image is annotated with a fine-grained attribute. There are eight attributes,
and each attribute is associated with a list of sub-classes. For instance, the attribute neckline design has 11 corresponding sub-classes such as round neckline and v neckline. We randomly split images into three sets with an 8:1:1 ratio, resulting in 144k/18k/18k images for training/validation/testing. Besides, for every epoch, we construct 100k triplets from the training set for model training. Concretely,
for a triplet with respect to a specific attribute, we randomly sample two images of the same corresponding sub-classes as the relevant pair and an image with differentsub-classes as the irrelevant one. For validation or test set, 3,600 images are randomly picked as query images. The remaining images, annotated with the same attribute, are used as the candidate images for retrieval.

\textbf{\noindent DeepFashion.}
DeepFashion~\cite{Liu_CVPR_16} is a large dataset that consists of four benchmarks for tasks in clothing: category and attribute prediction, in-shop clothes retrieval, fashion landmark detection, and consumer-to-shop clothes retrieval. In our experiments, we use the category and attribute prediction benchmark for the attribute-specific retrieval task. 

The category and attribute prediction benchmark contains
289,222 images with six attributes and 1,050 sub-classes, and
each image is annotated with several attributes. We randomly split the images into training, validation, and test sets by an 8:1:1 ratio and construct 100k triplets for training. For both the validation and test sets, images are split into query and candidate images at a 1:4 ratio.

\textbf{\noindent Zappos50k.}
Zappos50k~\cite{Yu_CVPR_14} is a large shoe dataset consisting of 50,025
images collected from the online shoe and clothing retailer,
Zappos.com. For ease of cross-paper comparison, we use the identical split provided by~\cite{Veit_CVPR_17}. Specifically, we use 70\%/10\%/20\% images for training/validation/testing. Each image is tagged with four attributes: the type of the shoes, the suggested gender of the shoes, the height of the shoes’ heels, and the closing mechanism of the shoes. For each attribute, 200k training,
20k validation, and 40k testing triplets are sampled for model training and evaluation.

\section{Experimental results}
\subsection{Experimental settings}

To assess the feasibility of the proposed model, we conduct evaluations on the attribute-specific fashion retrieval (ASFR) and triplet relation prediction (TR) tasks. Details of the definition of these two tasks and the experiments are given below.

\textbf{\noindent ASFR task.} When given a fashion image along with a specified attribute, the objective is to identify fashion images from the dataset that share the same attribute as the given reference image. We employ two datasets for this task: the FashionAI dataset~\cite{Ma_AAAI_20} and the DeepFashion dataset~\cite{Liu_CVPR_16}. Specifically, 1) The FashionAI dataset encompasses eight attributes, each featuring multiple values. These attributes can be categorized into design-related and length-related attributes. 2) In contrast, the DeepFashion dataset comprises five attributes, with approximately 200 values per attribute. These 13 attributes collectively pertain to distinct regions within images and encompass a wide spectrum of image features, ranging from low-level characteristics to high-level image comprehension. Therefore, experiments on these two datasets allow us to assess the proposed model's real-world applicability and effectiveness. Our evaluation metric is the Mean Average Precision (MAP), which is a widely recognized performance measure employed in various retrieval-related tasks~\cite{Ma_AAAI_20}. 

\textbf{\noindent TR task} With a triplet of $\{(I, I^{p}, I^{n})|a\}$ as input, it aims to predict whether the relevance degree of $(I,I^{p})$ surpasses that of $(I,I^{n})$ for the specified attribute $a$. The dataset used in this task is the Zappos50k dataset~\cite{Yu_CVPR_14}. We measure performance using prediction accuracy. 

The batch size, embedding size, learning rate, learning rate decay step size, and decay rate are 16, 1024, 0.0001, 3, and 0.9, respectively. The training epoch number is 50. The Figure~\ref{fig:compared_models} illustrates the conceptual frameworks of the models under comparison, as well as our own model.

\subsection{Results}

\subsubsection{Experimental results on the ASFR task}
We compare the proposed model with typical attention networks for fine-grained fashion retrieval task to show that our ways to improve the attribute-aware attention module and alleviate feature gap problem are effective. We do not compare our work with Yan {\it et al.}~\cite{Yan_ICME_22}, due to the absence of released source code, preventing us from re-implementing their approach and replicating the reported results. Tables~\ref{tab:Main_results_FashionAI} and~\ref{tab:Main_results_DeepFashion} present an overview of the comparison models' performance on the FashionAI and DeepFashion datasets, respectively. The tables display the results for various attribute types. It is evident that the proposed network consistently exhibits better performance than other models on both FashionAI and DeepFashion datasets.

Figure~\ref{fig:FashionAI_retri} presents the top-10 retrieval examples generated by the proposed model when query cloth and attribute are given. The accurate results in the top-10 retrieval examples serve as compelling evidence of the AG-MAN model's efficacy. 

We also show some ranking results. Given a query image and 10 candidate clothes, the model evaluates and ranks the candidates based on their similarity to the query image with respect to a specified attribute of interest. This process involves extracting and comparing detailed feature representations of both the query image and each candidate garment to determine their match level in terms of the query attribute. The candidates are then ordered from most to least similar, facilitating the identification of garments that best match the user's search criteria. As we can see from Figure~\ref{fig:ranking}, even when other models fail, the proposed method can produce an accurate ranking. It embeds the labeled similar items closer than the labeled different ones.

To gain deeper insights into our proposed model, we visualize the attention maps learned by the proposed ASA. As depicted in Figure~\ref{fig:Attenmap}, these attention maps exhibit notable characteristics. They tend to exhibit higher response values in regions relevant to the specified attribute, while diminishing responses in regions deemed irrelevant. This can demonstrate the ASA's capability to discern which areas are more pertinent for a specific attribute. It's worth noting that attention maps for attributes related to item length exhibit greater complexity compared to design-related attributes. In the case of length-related attributes, multiple regions often exhibit heightened responses. However, our model excels in accurately locating the starting and ending points of a fashion item, enabling it to make informed speculations about its length. Even in scenarios involving the most intricate global attributes, our model consistently outperforms in pinpointing the relevant regions.

\subsubsection{Experimental results on the TR task}
\begin{figure}[t]
  \centering
   \includegraphics[width=0.98\linewidth]{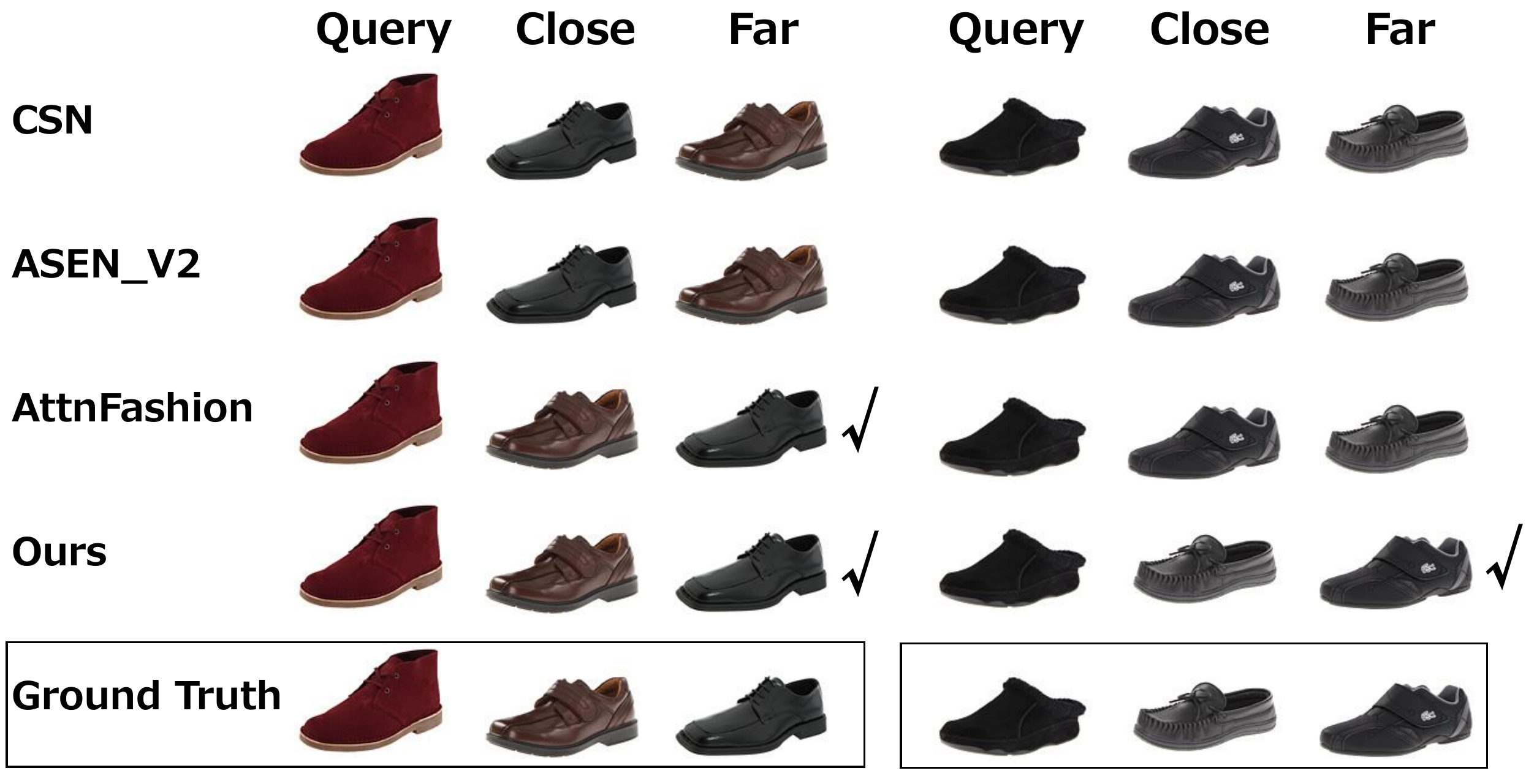}
   \caption{Visualization examples for triplet relation prediction. The `Close' column is the predicted closer item, while the `Far' column denotes the predicted farther item.}
   \label{fig:Zappos}
    \vspace{-0.2cm}
\end{figure}

Table~\ref{tab:Main_results_Zappos50k} presents the results for triplet relation prediction on the Zappos50k dataset. Out of the four methods for fine-grained fashion retrieval, our proposed model demonstrates superior performance compared to other methods. Figure~\ref{fig:Zappos} shows some triplet relation prediction examples. As we can see, the proposed method can accurately predict the relation of a triplet input while other models fail.

\begin{table}[t]
\caption{Triplet relation prediction on Zappos50k.}
\begin{center}
\resizebox{0.48\textwidth}{!}{
\begin{tabular}{lcc}
\toprule
\multirow{2}{*}{Methods}&\multicolumn{2}{c}{Zappos50k}\\
\cline{2-3}
  &Average loss $\downarrow$ & Prediction Accuracy(\%) $\uparrow$ \\ \midrule
 CSN~\cite{Veit_CVPR_17} & \underline{0.0393} &  \underline{93.14}  \\ \midrule
  ASENet\_V2~\cite{Ma_AAAI_20}  &0.0430 &  92.54\\ \midrule
   AttnFashion~\cite{Wan_ETAI_22}  & 0.0664 &   91.37 \\ \midrule
 Ours&  \textbf{0.0390} & \textbf{93.32}
  \\ \bottomrule
\end{tabular}}
\end{center}
\label{tab:Main_results_Zappos50k}
 \vspace{-0.2cm}
\end{table}

\begin{table*}[t]
\caption{Ablation studies on the FashionAI dataset.}
\begin{center}
\resizebox{1.0\textwidth}{!}{
\begin{tabular}{lcccccccccccc}
\toprule
\multirow{2}{*}{Methods}&\multirow{2}{*}{AGA} &\multirow{2}{*}{\thead{Hierarchical \\ feature}}&\multirow{2}{*}{\thead{Classification\\loss}} & \multicolumn{8}{c}{MAP for each attribute (Fashion AI)}& \multirow{2}{*}{MAP $\uparrow$}\\
\cline{5-12}
 & &  & &Skirt-length & Sleeve-length & Coat-length & Pant-length &Collar-design &Lapel-design &Neckline-design &Neck-design & \\
\midrule
\multirow{4}{*}{Ours} &\checkmark&-&-& 66.2107 & 55.3089  &  52.6219 &68.3924  & 71.2480 &  66.8022  & 62.1120   & 61.9079   &61.9679     \\
&\checkmark&\checkmark&-&65.1353  & 57.0038 &  53.4465 & 68.0217 & 72.2718 &67.4483   &62.6834   &61.9855   &  62.4804    \\

&\checkmark&-&\checkmark & 65.8052  &  55.4035 &  54.3290 & 68.0434 & 72.4414 &65.3819   &62.6148    & 62.5654   &   62.2951   \\
&\checkmark&\checkmark&\checkmark &65.5674  & 55.0840 &  55.2060& 68.9536 &73.7867 &68.1758   &63.1985   &62.0777   & 62.8788  \\ \bottomrule
\end{tabular}}
\end{center}
\label{tab:Ablation_FashionAI}
 \vspace{-0.2cm}
\end{table*}

\begin{table*}[t]
\caption{Ablation studies on the DeepFashion dataset.}
\begin{center}
\resizebox{0.90\textwidth}{!}{
\begin{tabular}{lccccccccc}
\toprule
\multirow{2}{*}{Methods}&\multirow{2}{*}{AGA} &\multirow{2}{*}{Hierarchical feature}&\multirow{2}{*}{Classification loss} & \multicolumn{5}{c}{MAP for each attribute (DeepFashion)}  & \multirow{2}{*}{MAP $\uparrow$}\\
\cline{5-9}
& &  & &Texture-related & Fabric-related & Shape-related & Pant-related
 &Style-related & \\
\midrule

\multirow{4}{*}{Ours} &\checkmark& -& - &  15.2951 &  7.1808 & 12.2927 &5.6646  &  3.6250 & 8.8069\\ 
 &\checkmark& \checkmark& - &  15.0432 & 7.2492 & 12.2757 &5.8363 &  3.8243 &8.8406\\ 
 &\checkmark& -&\checkmark &  15.2251 & 7.4224 & 12.5469 &6.0820 & 4.0149 &9.0512\\ 
 &\checkmark& \checkmark &\checkmark &  15.6624 & 7.2746 & 12.3805 &5.8935 & 3.7555 &8.9804\\
  \bottomrule
\end{tabular}}
\end{center}
\label{tab:Ablation_DeepFashion}
 \vspace{-0.2cm}
\end{table*}

\subsection{Ablation studies}

Ablation studies were carried out to evaluate the performance changes of the proposed AG-MAN when removing one of the following: hierarchical feature and classification branch. As we can see from Tables~\ref{tab:Ablation_FashionAI} and~\ref{tab:Ablation_DeepFashion}, both the hierarchical feature and classification branch can improve the model's performance on the two datasets considered in this paper. The optimal performance is achieved when both components are retained in the model, further underscoring their importance.

We also conduct experiments to evaluate the performance changes when removing ASA, SA, ACA, or CA from the proposed AGA module. The experimental results are shown in Tables~\ref{tab:Ablation_AGA_FashionAI} and~\ref{tab:Ablation_AGA_DeepFashion}. As we can see, when combining all of them, the best performance can be achieved.

\begin{table*}[t]
\caption{Ablation studies on the proposed AGA module (FashionAI dataset).}
\begin{center}
\resizebox{1.0\textwidth}{!}{
\begin{tabular}{lccccccccccccc}
\toprule
\multirow{2}{*}{Methods}&\multirow{2}{*}{ASA} &\multirow{2}{*}{SA}&\multirow{2}{*}{ACA}&\multirow{2}{*}{CA}  & \multicolumn{8}{c}{MAP for each attribute (Fashion AI)}& \multirow{2}{*}{MAP $\uparrow$}\\
\cline{6-13}
 & &  & & &Skirt-length & Sleeve-length & Coat-length & Pant-length &Collar-design &Lapel-design &Neckline-design &Neck-design & \\
\midrule
\multirow{5}{*}{Our AGA} &-&\checkmark&\checkmark&\checkmark& 59.3651  &  43.3860 &  43.5791 & 62.3133  &66.8545  & 54.7183   & 40.1253   & 54.1803  & 50.7971   \\ 
&\checkmark &- &\checkmark &\checkmark& 65.7707  &  53.4347 & 52.7763  & 66.7755  &72.3489  &   67.9203 &  62.6388  & 61.2233  & 61.6719   \\ 
&\checkmark  &\checkmark &- &\checkmark & 64.7258  &  55.3104 & 52.6036  & 68.3390  & 71.8768 &  65.7550  &  60.6618  & 62.1818  &  61.4976  \\ 
&\checkmark  &\checkmark &\checkmark &- &  65.6344 & 54.6032  & 52.5263  & 65.9612  &71.7295  &  67.6182  & 61.2087   &  61.8379 & 61.4086   \\ 
&\checkmark  &\checkmark &\checkmark &\checkmark &65.5674  & 55.0840 &  55.2060& 68.9536 &73.7867 &68.1758   &63.1985   &62.0777   & 62.8788  \\ \bottomrule
\end{tabular}}
\end{center}
\label{tab:Ablation_AGA_FashionAI}
 \vspace{-0.2cm}
\end{table*}

\begin{table*}[t]
\caption{Ablation studies on the proposed AGA module (DeepFashion dataset).}
\begin{center}
\resizebox{0.90\textwidth}{!}{
\begin{tabular}{lcccccccccc}
\toprule
\multirow{2}{*}{Methods}&\multirow{2}{*}{ASA} &\multirow{2}{*}{SA}&\multirow{2}{*}{ACA}&\multirow{2}{*}{CA} & \multicolumn{5}{c}{MAP for each attribute (DeepFashion)} & \multirow{2}{*}{MAP $\uparrow$}\\
\cline{6-10}
& &  &  & &Texture-related & Fabric-related & Shape-related & Pant-related
 &Style-related & \\
\midrule
\multirow{5}{*}{Our AGA}  &-&\checkmark&\checkmark&\checkmark& 14.6402  &  6.3866  & 10.4550  & 4.7022  &  3.3938  & 7.8930   \\ 
&\checkmark &- &\checkmark &\checkmark&  15.4613  & 7.0831  & 12.2323  & 5.4195  &  3.5369 &  8.7390\\ 
&\checkmark  &\checkmark &- &\checkmark &  15.1934  & 6.8899  &  11.1476 &  5.4773 &3.6428   & 8.4511 \\ 
&\checkmark  &\checkmark &\checkmark &-&  15.1324  & 7.1874  &  11.6915 & 5.6828  & 3.6776  &8.6716  \\ 
&\checkmark  &\checkmark &\checkmark &\checkmark&  15.6624 & 7.2746 & 12.3805 &5.8935 & 3.7555 &8.9804\\
  \bottomrule
\end{tabular}}
\end{center}
\label{tab:Ablation_AGA_DeepFashion}
 \vspace{-0.2cm}
\end{table*}

\subsection{Other experiments}

To verify that our AGA is an improvement of the one in~\cite{Ma_AAAI_20}, and better than the atrribute-aware attention modules proposed in other methods, we conduct experiments on both the FashionAI and DeepFashion datasets. As we can see from Tables~\ref{tab:OtherExp_FashionAI} and~\ref{tab:OtherExp_DeepFashion}, the proposed AGA is more effective compared with exiting attribute-specific attention modules. It shows that by adding the SA and CA module, they obtain a good coupling effect. 

\begin{table*}[t]
\caption{Comparison of the proposed AGA and exiting attribute-aware attention modules (FashionAI).}
\begin{center}
\resizebox{0.98\textwidth}{!}{
\begin{tabular}{lccccccccc}
\toprule
\multirow{2}{*}{Methods}& \multicolumn{8}{c}{MAP for each attribute (Fashion AI)} & \multirow{2}{*}{MAP $\uparrow$}\\
\cline{2-9}
  &Skirt-length & Sleeve-length & Coat-length & Pant-length &Collar-design &Lapel-design &Neckline-design &Neck-design & \\
\midrule
 ASENet\_V2~\cite{Ma_AAAI_20} &\underline{64.5731} &\underline{54.9630} &\underline{51.7645} & \underline{64.4973} &\textbf{71.9256}& \underline{66.7201}& \underline{60.2929} &\underline{60.8307}&\underline{60.7551}  \\ \midrule
 AttnFashion~\cite{Wan_ETAI_22} & 61.9057&45.8526 &45.8116 & 60.6013&70.4638 &62.7362 &52.0484 &59.5861 &55.3646   \\ \midrule
Ours (only AGA) & \textbf{66.2107} & \textbf{55.3089}  &  \textbf{52.6219} & \textbf{68.3924}  & \underline{71.2480} &  \textbf{66.8022}  & \textbf{62.1120}   & \textbf{61.9079}   &\textbf{61.9679}   \\ \bottomrule
\end{tabular}}
\end{center}
\label{tab:OtherExp_FashionAI}
 \vspace{-0.2cm}
\end{table*}

\begin{table*}[t]
\caption{Comparison of the AGA and other attribute-aware attention modules (DeepFashion).}
\begin{center}
\resizebox{0.8\textwidth}{!}{
\begin{tabular}{lcccccc}
\toprule
\multirow{2}{*}{Methods}& \multicolumn{5}{c}{MAP for each attribute (DeepFashion)} & \multirow{2}{*}{MAP $\uparrow$}\\
\cline{2-6}
  &Texture-related & Fabric-related & Shape-related & Pant-related
 &Style-related & \\
\midrule
  ASENet\_V2~\cite{Ma_AAAI_20} &\textbf{15.5168}
  &\textbf{7.1855} &\underline{11.5066} &\underline{5.5220} &\textbf{3.6514} &  \underline{8.6704}\\ \midrule
 AttnFashion~\cite{Wan_ETAI_22} &11.8423  &5.7248 &8.6115 &4.0904 &3.2329 & 6.8060 \\ \midrule
  Our AGA &  \underline{15.2951} &  \underline{7.1808} & \textbf{12.2927} &\textbf{5.6646}  &  \underline{3.6250} & \textbf{8.8069} \\ 
 \bottomrule
\end{tabular}}
\end{center}
\label{tab:OtherExp_DeepFashion}
 \vspace{-0.2cm}
\end{table*}

\subsection{Failure analysis}
\begin{figure}[t]
  \centering
   \includegraphics[width=0.8\linewidth]{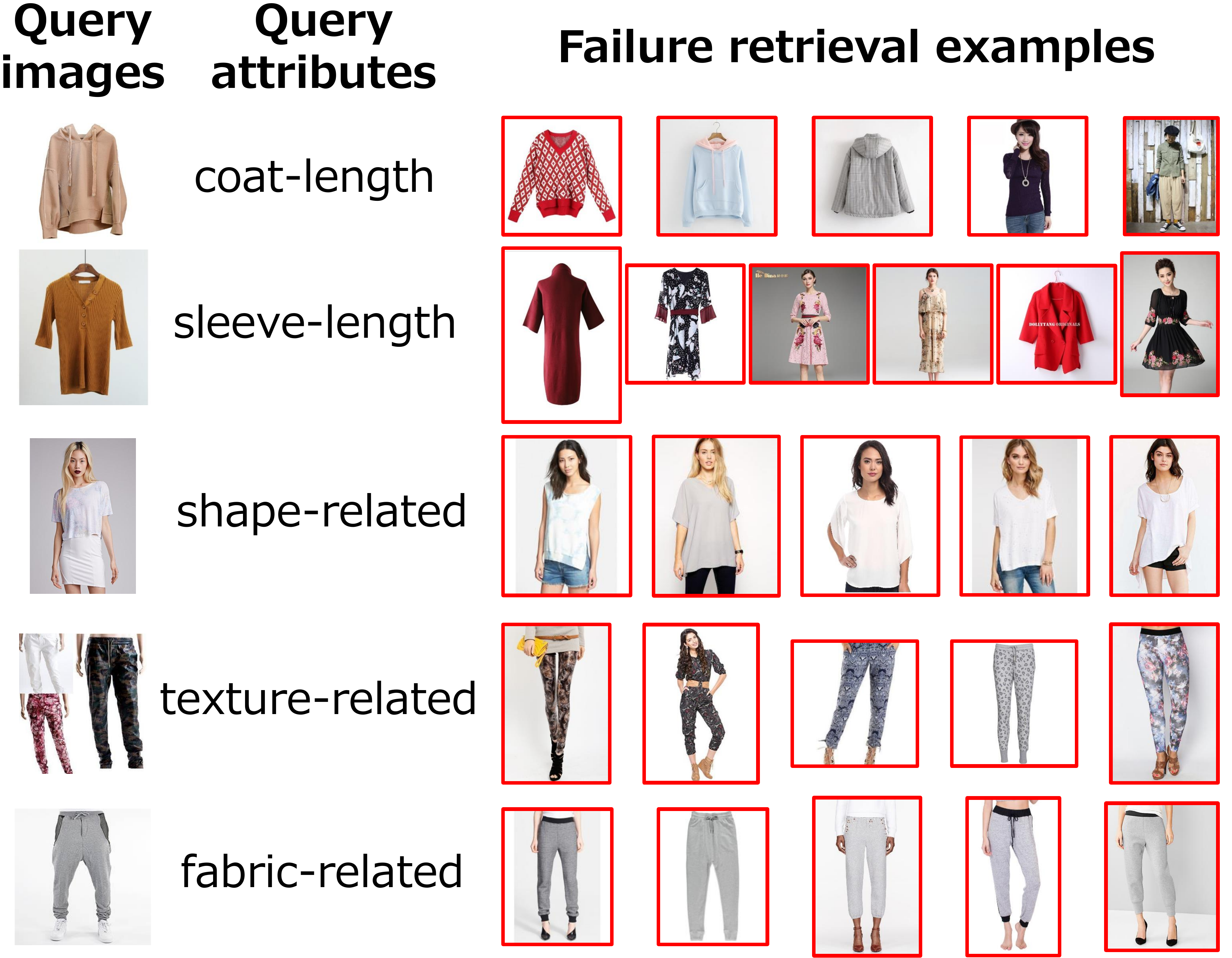}
   \caption{Some retrieval failures of the proposed method.}
   \label{fig:retrieval_failure}
    \vspace{-0.2cm}
\end{figure}
Figure~\ref{fig:retrieval_failure} shows some failure cases of our model. First, our model may incorrectly evaluate on some items that share very small differences with the query image, as seen in texture-related fashion retrieval in the figure. For certain attributes related to length, some fashion images are labeled with varying lengths that are so subtle that even human observers find it challenging to discern the differences accurately. Moreover, real-world datasets include pure item and try-on images that are taken from different viewpoints, making it difficult to distinguish the subtle difference in length between these two kinds of images. Finally, the images with occlusions also add some difficulty. In the future, we aim to address these issues mentioned above.

\section{Limitations and future works}
The proposed method is not scalable. When new attributes come, we need to train the whole model again. Scalable fine-grained fashion retrieval by designing novel continual learning and Large Language Model prompt techniques would be a promising research direction.

\section{Conclusions}
This paper introduced a novel approach called the AG-MAN to address the feature gap problem and proposed a  better attribute-aware attention module. Concretely, the AG-MAN modified the pre-trained CNN backbone to capture multi-level feature representations for addressing the feature gap problem. We also proposed a method to disturb the object-centric feature learning to further alleviate this problem. Moreover, we proposed a more powerful attribute-guided attention module for extracting more discriminative representations. AG-MAN exceled in achieving precise attribute positioning and extracting highly distinctive features, while being guided by the specified attribute. Extensive experiments with the FashionAI, DeepFashion, and Zappos50k datasets showed that our AG-MAN surpasses existing attention networks for fine-grained fashion retrieval. Our AG-MAN exhibits significant promise for application in various attribute-guided recognition tasks.

\section{Acknowledgments}
This paper is partially financially supported by Institute for AI and Beyond of The University of Tokyo.
{
    \small
    \bibliographystyle{ieeenat_fullname}
    \bibliography{main}
}


\end{document}